\pdfoutput=1

\def\year{2020}\relax
\documentclass[letterpaper]{article} % DO NOT CHANGE THIS
\usepackage{aaai20}  % DO NOT CHANGE THIS
\usepackage{times}  % DO NOT CHANGE THIS
\usepackage{helvet} % DO NOT CHANGE THIS
\usepackage{courier}  % DO NOT CHANGE THIS
\usepackage[hyphens]{url}  % DO NOT CHANGE THIS
\usepackage{graphicx} % DO NOT CHANGE THIS
\usepackage{graphicx}  % DO NOT CHANGE THIS
\usepackage[utf8]{inputenc} % allow utf-8 input
\usepackage{url}            % simple URL typesetting
\usepackage{booktabs}       % professional-quality tables
\usepackage{amsfonts}       % blackboard math symbols
\usepackage{nicefrac}       % compact symbols for 1/2, etc.
\usepackage{microtype}      % microtypography

\usepackage{graphicx}
\usepackage{subfigure}

\usepackage{amsmath,tikz,bm}
\usepackage{adjustbox}      
\usepackage{amssymb,enumitem,filecontents,multirow}
\usepackage{algorithmic,algorithm}
\usepackage{array,rotating,graphicx}
\usepackage{xspace}

\newcommand{\bfx}{\mathbf{x}}
\newcommand{\bfX}{\mathbf{X}}
\newcommand{\bfz}{\mathbf{z}}
\newcommand{\bfa}{\mathbf{a}}
\newcommand{\bfc}{\mathbf{c}}
\newcommand{\bff}{\mathbf{f}}
\newcommand{\bfh}{\mathbf{h}}
\newcommand{\bfq}{\mathbf{q}}

\newcommand{\bfalpha}{\bm{\alpha}}
\newcommand{\bfv}{\mathbf{v}}
\newcommand{\calT}{\mathcal{T}}
\newcommand{\calE}{\mathcal{E}}
\newcommand{\calV}{\mathcal{V}}
\newcommand{\Epsilon}{E}

\newcommand{\mbf}[1]{{\boldsymbol{\mathbf{#1}}}}
\renewcommand{\bm}{\mbf}

\newcommand{\mname}{\texttt{CORE}\xspace }
\newcommand{\oov}{{OOI}\xspace }

\newcolumntype{L}[1]{>{\raggedright\arraybackslash}p{#1}}
\newcolumntype{R}[1]{>{\raggedleft\arraybackslash}p{#1}}
\newcolumntype{C}[1]{>{\centering\let\newline\\\arraybackslash\hspace{0pt}}m{#1}}
\newcolumntype{?}{!{\vrule width 1pt}}
\makeatletter
\newcommand{\thickhline}{%
    \noalign {\ifnum 0=`}\fi \hrule height 1pt
    \futurelet \reserved@a \@xhline
}
\newcolumntype{"}{@{\hskip\tabcolsep\vrule width 1pt\hskip\tabcolsep}}
\makeatother
\title{\mname: Automatic Molecule Optimization Using Copy \& Refine Strategy }
\author{
Tianfan Fu,\textsuperscript{\rm 1}
Cao Xiao,\textsuperscript{\rm 2}
Jimeng Sun\textsuperscript{\rm 1}\\
\textsuperscript{\rm 1} College of Computing, Georgia Institute of Technology, Atlanta, USA \\
\textsuperscript{\rm 2}Analytics Center of Excellence, IQVIA, Cambridge, USA
 \\tfu42@gatech.edu,
 cao.xiao@iqvia.com,
 jsun@cc.gatech.edu.
}

\begin{document}

\maketitle

\begin{abstract}
Molecule optimization is about generating molecule $Y$ with more desirable properties based on an input molecule $X$. 
The state-of-the-art approaches partition the molecules into a large set of substructures $S$ and  grow the new molecule structure by iteratively predicting which substructure from $S$ to add. However, since the set of available substructures $S$ is large, such an iterative prediction task is often inaccurate especially for substructures that are infrequent in the training data. To address this challenge, we propose a new generating strategy called ``\texttt{Co}py\&\texttt{Re}fine'' (\mname), where at each step the generator first decides whether to copy an existing substructure from input $X$ or to generate a new substructure, then the most promising substructure  will be added to the new molecule. Combining together with scaffolding tree generation and adversarial training, \mname can significantly improve several latest molecule optimization methods in various measures including drug likeness (QED), dopamine receptor (DRD2) and penalized LogP. We tested \mname and baselines using the ZINC database and \mname obtained up to 11\% and 21\% relatively improvement over the baselines on success rate on the complete test set and the subset with infrequent substructures, respectively. 
\end{abstract}

\section{Introduction}
Designing molecules or chemical compounds with desired properties is a fundamental task in drug discovery. 
Since the number of drug-like molecule is large as estimated between $10^{23}$ and $10^{60}$~\cite{polishchuk2013estimation}, traditional methods such high throughput screening (HTS) is not scalable. One particular task in drug discovery is called {\it lead optimization}, where after a drug candidate (a {\it hit}) is identified via HTS, enhanced similar candidates are created and tested in order to find a lead compound with better properties than the original hit. 
To model lead optimization as a machine learning problem, the training data involves a set of paired molecules that map input molecule $X$ to target molecule $Y$ with more desirable properties. The goal is to learn a generating model that can produce target molecules with better properties from an input molecule.

Automatic molecular generation algorithms have made great progress in recent years thanks to the successful use of deep generative models~\cite{jin2018junction,liu2018constrained,zhou2019optimization,jin2019learning}. Early ones cast the molecular generation as a sequence generation problem since a molecule can be represented by a SMILES string\footnote{The Simplified Molecular-Input Line-Entry System (SMILES) is a specification in the form of a line notation for describing the structure of chemical species using short ASCII strings.}~\cite{kusner2017grammar,dai2018syntax,gomez2018automatic}. 
 However, many of those algorithms generate many invalid SMILES strings which do not correspond to any valid molecules. 
 
 \noindent{\bf Progress:} Recent years graph generation methods have been proposed to bypass the need to produce a valid SMILES string by directly generating molecular graphs\cite{jin2018junction,jin2019learning,NIPS2018_7942}.
%graph-to-Graph~\cite{jin2019learning} achieve state-of-the-art performance in generating a molecule with certain chemical property. 
These graph based approaches reformulate the molecular generation task as a graph-to-graph translation problem, which  avoid the need of generating SMILES strings. Their key strategy is to partition the input molecule graph into a scaffolding tree of substructures (e.g., rings, atoms, and bonds), and to learn to generate such a tree. All possible tree nodes lead to a large set of substructures, e.g., around 800 unique substructures in the ZINC database~\cite{sterling2015zinc}.

\noindent{\bf Challenge: } However, the graph generation methods still exhibit  undesirable behaviors such as generating inaccurate substructures because the set of all possible substructures is large, especially for infrequent substructures. 
In each generating step, the model has to decide which substructure to add from a large set of possible substructures. 
%At the same time, the size of whole vocabulary
%This ``generative'' process make the prediction very challenging. 
On the other hand, from real data we observe
\begin{itemize}
    \item {\bf Stable principle:} Vast majority of substructures in target molecules are from the input molecule. The first row of Table~\ref{table:oov} shows about and over 80\% substructures are from the input molecule in four datasets/tasks. 
    \item {\bf Novelty principle:} New substructures are present in most target molecules.  The second row of Table~\ref{table:oov} shows that for 80\% target molecules have new substructures compared with their corresponding input molecules. 
    % \item {\bf Search Space Difference:} 
    % From the third and fourth row of Table~\ref{table:oov}, each molecule have 14 substructures in average while the global set of substructures is at least 780. 
\end{itemize}

% find that most of the substructures in the target molecule are already existing in the input molecule. 
% We consider an ``extractive'' method that copy some substructures from input molecule when generating molecule so that we can prune the search space and don't have to do the prediction over the whole vocabulary every time. 
% In addition, we know from another empirical observation that in most of the data pairs, the target molecule contains some new substructures that don't occur in input molecule. 
% So a purely ``extractive'' way is not enough. 

\begin{table}[tb]
\centering
\caption{\footnotesize Comparison between input and target molecules on 4 datasets/tasks. {\bf Stable principle}: Row 1 shows the percentage of original substructures in the target molecule, which is about 80\% or more and indicates many original substructures are kept in the newly generated targets. {\bf Novelty principle}: Row 2 shows the percentage of targets have any new substructures that do not belong to the input molecule, which is also high and indicates the need for including new substructures in the targets.
Row 3 lists the number of all the substructure, i.e., $\vert S\vert$ and Row 4 lists the average substructures for molecules. }
\label{table:oov}
\begin{tabular}{c|c|c|c|c}
\toprule[1pt]
 & DRD2 & LogP04 & LogP06 & QED \\ \hline 
\% of original& 80.42\% & 79.47\% & 88.90\% & 83.32\% \\ 
\% of novel & 86.40\% &  84.06\% & 70.14\% & 80.84\% \\ 
\# substructures & 967 & 785 & 780 & 780 \\ 
Molecule size & 13.85 & 14.30 & 14.65 & 14.99 \\
\bottomrule[1pt]
\end{tabular}
\end{table}

Based on these observations, we propose a new strategy for molecular optimization called {\it Copy \& Refine} (\mname). The key idea is at each generating step, \mname will decide whether to copy a substructure from the input molecule ({\it Copy}) or sample a novel substructure from the entire space of substructures ({\it Refine}).  

% Meanwhile, we are also inspired by pointer network, which copy large chunks of text from source document in Sequence-to-Sequence model and widely used in natural language processing community~\cite{vinyals2015pointer,gu2016incorporating,see2017get}. 
% Concretely, to solve these issues, we design a finer-grained architecture to allow (1) copy some substructures from input molecule; (2) generate a new substructures from the whole vocabulary. 

We compare \mname with different baselines and demonstrate significant performance gain in several metrics including drug likeness (QED), dopamine receptor (DRD2) and penalized LogP.  
We tested \mname and baselines using the ZINC database and \mname achieved up to 11\% and 21\% relatively improvement over the baselines on success rate on the complete test set and the subset with infrequent substructures, respectively.

The rest of paper is organized as follows. We briefly review related work on molecule generation. Then we describe \mname  method. 
Finally, we show empirical studies and conclude our paper.

\section{Related Work}
We review  related works in molecule generation. %using sequence based methods and graph based methods. 
%research lines in automatic molecule generation. First is based on sequence representation of molecule. Second is graph representation. 
\paragraph{Sequence-based methods} 
One research line is to formulate molecular generation as a sequence based problem~\cite{dai2018syntax,kusner2017grammar,zhou2017private,gomez2018automatic,HongXMLS19,huang2019}.
Most of these methods are based on the simplified molecular-input line-entry system (SMILES), a line notation describing the molecular structure using short ASCII strings~\cite{weininger1988smiles}.  
Character Variational Auto-Encoder (C-VAE) generates SMILES string character-by-character \cite{gomez2018automatic}. 
Grammar VAE (G-VAE) generates SMILES following syntactic constraints given
by a context-free grammar~\cite{kusner2017grammar}.  
Syntax-directed VAE (SD-VAE) that incorporates both syntactic
and semantic constraints of SMILES via attribute grammar~\cite{dai2018syntax}. 
Reinforcement learning (RL) is also used to generate SMILES strings~\cite{Popova2018-ws,Olivecrona2017-ry}.
However, many generated SMILES strings using these methods are not even syntactically valid which lead to inferior results.

\paragraph{Graph-based methods} 
Another research line is to directly generate molecule graphs. %\cx{Molecules can be represented as undirected graphs with atoms as nodes and bonds as edges. As sequence based approach is very brittle since small changes to the sequence may completely violate chemical validity, they need carefully designed constraints to ensure validity. As a contrast, graph is a more natural way to represent molecule and to capture the structural information among atoms of a molecule }
Comparing sequence based method, graph-based methods bypass the need of generating valid sequences (e.g., SMILES strings) all together. As a result, all the generated molecules are valid and often with improved properties. 
The original idea of \cite{jin2018junction} is to eliminate cycles in a molecular graph by representing the graph as a scaffolding tree where nodes are substructures such as rings and atoms. Then a two-level generating function is used to create such a tree then decode the tree into a new molecular graph. 
Recently  another enhancement is produced called graph-to-graph translation model~\cite{jin2019learning}, which extends the junction variational autoencoder via attention mechanism and generative adversarial networks. It is able to translate the current molecule to a similar molecule with prespecified desired property (e.g., drug-likeness). Also a RL based method with graph convolutional policy network has been proposed to generate molecular graphs with desirable properties~\cite{You2018-xh}.
%\js{check if the general statement next is applicable to the RL method \cite{You2018-xh}, if not we need to have a separate limitation for \cite{You2018-xh}}
However, these graph based models require to iteratively predict the best substructure to add from a large set of possible choices, which is often inaccurate especially for the rare substructures. 
We overcome this limitation using a hybrid strategy which involves copy from the input molecule then refine it by selectively adding new substructures. 
%Our method is modified based Graph-to-Graph model \ari{tell this in the next section not here.}. 

\begin{table}[tb]
\centering
\caption{Important notations used in the paper.}
\label{table:notation}
\resizebox{\columnwidth}{!}{
\begin{tabular}{c|l}
\toprule[1pt]
Notations & short explanation \\ 
\hline 
$(X,Y)$ & input-target molecule pair \\ 
$S$ & substructure set $S$ (a.k.a. vocabulary) \\ 
 $V / \Epsilon$ & set of vertex(atom) / edge(bond)  \\ 
 $G = (V, \Epsilon)$ & molecular graph \\
 $\calT_G = (\calV, \calE)$  & scaffolding tree of graph $G$ \\ 
 $N(v)$ & set of neighbor nodes of vertex $v$  \\
 $\bff_v / \bff_{uv}$ & feature vector for node $v$ / edge $(u,v)$ \\
 $\bfv_{uv}^{(t)}$ & message for edge $(u,v)$ at the $t$-th iteration \\
 $\bfx_i^G / \bfx_i^{\calT}$ & embedding of node $i$ in  $G$ /  $\calT$, \\
 $\bfX^G$ & the set of node embedding $\bfX^G =\{\bfx_1^G,\cdots\}$ \\ 
 $\bfh_{i_t, j_t}$ &  message vector for edge $(i_t, j_t)$ \\ 
$\bfz_G$ & Embedding of Graph $G$ \\
$p_t^{\text{topo}}$ &  topological prediction score \\
$\bfq_t^{\text{sub}} / \tilde{\bfq}_t^{\text{sub}} $ & pred. dist. over all  substructures  \\
$g_i(\cdot), \  i=1,\cdots,6 $ & parameterized neural networks \\
\bottomrule[1pt]
\end{tabular}}
\end{table}

\section{Method}
\label{sec:method} 
In this section, we first describe overall framework of \mname which shares the same foundation as \cite{jin2019learning} . Then we present model enhancement that \mname introduces namely the copy \& refine strategy. 
We list some important notations and their short explanations in Table~\ref{table:notation}. 
%Both of them can be applied on graph-to-graph model independently and they can be jointly used. 

\begin{figure*}[h!]
\includegraphics[width=2.0\columnwidth]{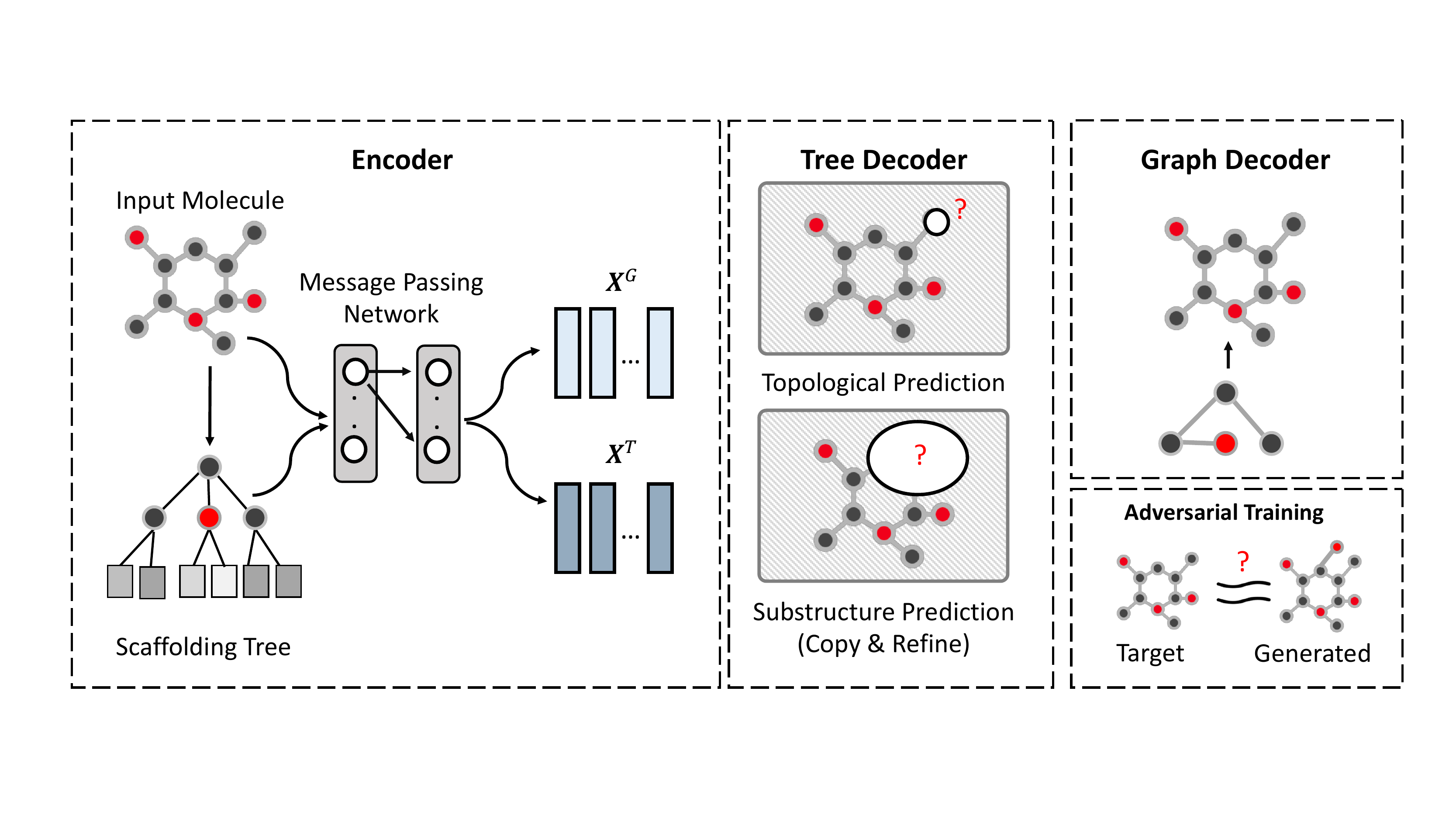}
%\vskip -1.5em
 \caption{Pipeline for Graph-to-Graph Model. 
Encoder include both graph and scaffolding tree levels.
Decoding mainly split two parts scaffolding tree decoder and graph decoder.
scaffolding tree decoder generate molecule in greedy manner using Depth First Search with topological and substructure prediction on each node. 
To assemble the node of scaffolding tree into the molecule, graph decoder enumerates all possible combinations. }
\label{fig:pipeline}
\end{figure*}

\subsection{Overview}
Graph-to-graph model involves two important structures: \begin{itemize}
    \item  {\bf molecular graph} $G$ is the graph structure for a molecule; 
    \item  {\bf scaffolding tree}  $\calT_G$ (also referred as scaffolding trees in \cite{jin2019learning}) is the skeleton of the molecular graph $G$ by partitioning the original graph into  substructures (subgraphs), and connecting those substructures into a tree.
\end{itemize}  

Given a molecule pair (input $X$ and target $Y$), we first train encoders to embed both $G$ and $\calT_G$ for input $X$ into vector representations via a message passing algorithm on graphs (or trees).  
%Then a tree-based recurrent neural network (tree-RNN) is trained on $Y$ and the embedding from $X$. 
Finally two-level decoders are introduced to create a new scaffold tree and the corresponding molecular graph. Our main methodology contribution lies in the decoder module where we propose a copy \& refine strategy to create novel but stable molecules from input molecules. The model is trained with a set of molecule pairs $(X,Y)$, where $Y$ is a target molecule that is optimized with better chemical properties based on the input molecule $X$. %$X$ is the input molecule and $Y$ is the target molecule in training procedure. 
%Following~\cite{jin2018junction}, 
%Each molecule are built from subgraphs, which collectively form a large set of valid chemical substructures. 
%To represent the scaffold tree \ari{It is not clear what  scaffold structure is?  }, it uses scaffolding tree formed by clusters (clusters of atoms).  
% Similar with Sequence-to-Sequence model, Graph-to-Graph model also contains encoder and decoder module. 
% Both encoder and decoder are decomposed into two parts: scaffolding tree and graph. 
% During decoding procedure, it \ari{it is not clear what is refers to?} adopts a coarse-to-fine manner. That is, first generating the scaffolding trees and then combining nodes in trees into molecule. 
% We start from the encoder. 

\subsection{Encoder}
To construct cycly-free structures, scaffolding tree $\calT_G$ is generated via contracting certain vertices of $G$ into a single node. 
By viewing scaffolding tree as graph, both input molecular graphs and scaffolding trees can be encoded via graph Message Passing Networks (MPN)~\cite{dai2016discriminative,jin2018junction}. 
The encoder yields an embedding vector for each node in either scaffolding tree or the input molecular graph. 
More formally, on node level $\bff_v$ denotes the feature vector for node $v$. 
For atoms, $\bff_v$ includes the atom type, valence, and other atomic properties. 
For nodes in the scaffolding tree representing substructures, $\bff_v$ is a one-hot vector indicating its substructure index. %\js{add concrete examples of a feature vector for an atom say carbon, and an example for a particular substructure}  
On the other hand, on edge level, $\bff_{uv}$ feature vector for edge $(u,v) \in \Epsilon$. 
$N(v)$ denotes the set of neighbor nodes for node $v$. 
$\bfv_{uv}$ and $\bfv_{vu}$ are the hidden variables that represent the message from node $u$ to $v$ and vice versa.  
They are iteratively updated via a fully-connected neural network $g_1(\cdot)$:
\begin{equation}
\begin{aligned}
\bfv_{uv}^{(t)} = g_1\big(\bff_{u}, \bff_{uv}, \sum_{w \in N(u)\backslash v} \bfv_{wu}^{(t-1)} \big),
\end{aligned}
\end{equation}
where $\bfv_{uv}^{(t)}$ is the message vector at the $t$-th iteration, whose initialization is $\bfv_{uv}^{(0)} = \mathbf{0}$. 
After $T$ steps of iteration, another network $g_2(\cdot)$ is used to aggregate these messages. Each vertex has a latent vector as
\begin{equation}
\begin{aligned}
\bfx_{u} = g_2\big(\bff_u, \sum_{v \in N(v)} \bfv_{vu}^{(T)} \big),
\end{aligned}
\end{equation}
where $g_2(\cdot)$ is another fully-connected neural network.

In summary, the encoder module yield embedding vectors for nodes in graph $G$ and scaffolding tree $\calT_G$, denoted $\bfX^{G}  = \{\bfx_{1}^{G}, \bfx_{2}^{G}, \cdots \}$ and $\bfX^{\calT_G}  = \{\bfx_{1}^{\calT_G}, \bfx_{2}^{\calT_G}, \cdots \}$, respectively. 

\subsection{Decoder}
Once the embedding vectors are constructed, decoder can also be divided into two phases in coarse-to-fine manner: (a) tree decoder; (b) graph decoder. We firstly discuss scaffolding tree decoder. Our method improve the tree decoder in \cite{jin2019learning}, so we describe the enhancement in detail.

% \begin{figure*}
%   \includegraphics[width=\linewidth]{figure/molecule1.pdf}
%   \caption{Graph-to-Graph Architecture}
%   \label{fig:graph2graph}
% \end{figure*}

\subsubsection{A. Tree decoder}
The objective of the scaffolding tree decoder is to generate a new scaffolding tree from the embeddings. 
The overall idea is to generate one substructure at a time from an empty tree, and at each time we decide whether to expand the current node or backtrack to its parent ({\it topological prediction}) and which to add ({\it substructure prediction}). The generation will terminate once the condition to backtrack from the root is reached.

%\js{what is the terminating condition? I think their original paper didn't see to specify, we should add.}
More specifically the tree decoder has two prediction tasks: 
\begin{itemize}
    \item {\bf Topological prediction:} When the decoder visit the node $i_t$, the model has to make a binary prediction on either ``expanding a new node'' or ``backtracking to the parent node of $i_t$''. 
    \item {\bf Substructure prediction:} If the decoder decides to expand, we have to select which substructure to add by either copying from original input or selecting from the global set of substructures. 
\end{itemize} 

\noindent{\bf Topological prediction:}
The idea is to first enhance the embedding for node $i_t$ via a tree-based RNN~ \cite{jin2018junction}, then use the enhanced embedding to predict whether to expand or backtrack. 
Given scaffolding tree $\calT_G = (\calV, \calE)$, the tree decoder uses the tree based RNN with attention mechanism to further improve embedding information learned from the original message-passing embeddings $X^\calT$.  
Since RNN works on a sequence, the tree converts into a sequence of nodes and edges via depth-first search.
Specifically, let $\tilde{\calE} = \{(i_1, j_1), (i_2, j_2), \cdots, (i_m, j_m) \}$ be the edges traversed in depth first search, each edge is visited twice in both directions, so we have $m = \vert \tilde{\calE} \vert = 2\vert\calE \vert$.   
Suppose $\tilde{\calE}_t $ is the first $t$ edges in $\tilde{\calE}$, message vector $\bfh_{i_t, j_t}$ is updated as:
\begin{equation}
\begin{aligned}
\bfh_{i_t, j_t} = \text{GRU}(\bff_{i_t}, \{\bfh_{k,i_t}\}_{(k,i_t) \in \tilde{\calE}_t, k \neq j_t}).
\end{aligned}
\end{equation}

The probability whether to expand or backtrack at node $i_t$ is computed via aggregating the embeddings $\bfX^{\calT} $, $\bfX^{G}$ and the current state $\bff_{i_t}$, $\sum_{(k,i_t)\in \tilde{\calE}_t} \bfh_{k,i_t}$ using a neural network $g_3(\cdot)$:  
\begin{equation}
\begin{aligned}
p_t^{\text{topo}} &= g_3(\bff_{i_t}, \sum_{(k,i_t)\in \tilde{\calE}_t} \bfh_{k,i_t}, \bfX^{\calT},  \bfX^{G}), \\  & \text{where}\ t=1,\cdots,m. 
\end{aligned}
\end{equation}
Concretely, firstly compute context vector $\bfc^{\text{topo}}_t$ using attention mechanism\footnote{same procedure as Equation~\eqref{eqn:attention} and \eqref{eqn:attentionweight}, but different parameters}, then concatenate $\bfc_t^{\text{topo}}$ and $\bff_{i_t}$, followed by a fully connected network with sigmoid activation.

\noindent{\bf Substructure prediction:} Once the node expansion is decided, we have to find what substructures to add.
Empirically we observe this step is most challenging one as it leads to largest error rate. For example, during training procedure of QED dataset, topological prediction and graph decoding (mentioned later) can achieve 99\%+ and 98\%+ classification accuracy, respectively. 
In contrast, substructure prediction can achieve at most 90\% accuracy, which is much lower.  We design \mname strategy to enhance this part. 
The idea is every time after expanding a new node, the model have to predict its substructure from all the substructures in vocabulary. 
First we use attention mechanism to compute context vector based on current message vector $\bfh_{i_t,j_t} $ and node embedding $\bfX^{\calT}, \bfX^{G} $:
\begin{equation}
\label{eqn:attention}
\begin{aligned}
\bfc_t^{\text{sub}} &= \text{Attention}(\bfh_{i_t,j_t}, \bfX^{\calT}, \bfX^{G}), \\
\end{aligned}
\end{equation}
Specifically, we firstly compute attention weight via
\begin{equation}
\label{eqn:attentionweight}
\begin{aligned}
\bfalpha_j^{\calT} & = g_4(\bfh_{k,i_t}, \bfx_j^{\calT}),\ \ \bfalpha_i^\calT \in \mathbb{R}, \\ 
[\bfalpha_1^{\calT}, \bfalpha_2^{\calT}, \cdots] &= \text{Softmax}([\bfalpha_1^{\calT}, \bfalpha_2^{\calT}, \cdots]),
\end{aligned}
\end{equation}
where $g_4(\cdot)$ is dot-product function~\cite{vaswani2017attention}. $\{\bfalpha^G\}$ are generated in same way. 
Then context vector is generated via concatenating tree-level and graph-level context vector
\begin{equation}
\label{eqn:attention2}
\begin{aligned}
\bfc_t^{\text{sub}} &= \Big[\sum_i \bfalpha_i^{\calT} \bfx_{i}^{\calT}, \sum_j \bfalpha_j^{G} \bfx_j^{G} \Big].
\end{aligned}
\end{equation}

\begin{table*}[tb]
\small 
\centering
\caption{Statistics of 4 datasets, DRD2, QED, LogP04 and LogP06. ``Avg \# Substructures'' is the average number of substructures per molecule. For each test sample, we generate 20 molecules using different random seeds. ``\# Infreq Test'' is the number of test samples that contain at least one infrequent substructure. }
\label{table:statistics}
\begin{tabular}{c|c|c|c|c|c|c}
\toprule[1pt]
Dataset & \# Training Pairs & \# Valid Pairs & \# Test ($\times 20$) & \# Infreq Test ($\times 20$) & Vocab Size, $\vert S\vert$ & Avg \# Substructures \\ \hline 
DRD2 & 32,404 & 2,000 & 1,000 & 405 & 967 & 13.85 \\  
QED  & 84,306 & 4,000 & 800 & 294 & 780 &  14.99 \\  
LogP04 & 94909 & 5,000 & 800 & 287 & 785 & 14.30 \\  
LogP06 & 71,284 & 4,000 & 800 & 250 & 780 & 14.65 \\  
\bottomrule[1pt]
\end{tabular}
\end{table*}

Then based on attention vector $\bfc_t^{\text{sub}}$ and message vector $\bfh_{i_t, j_t}$, $g_5(\cdot)$, a fully-connected neural network with softmax activation, is added to predict the substructure: 
\begin{equation}
\label{eqn:jtdecoder}
\begin{aligned}
\bfq^{\text{sub}}_t &= g_5(\bfh_{i_t, j_t}, \bfc_t^{\text{sub}}), 
\end{aligned}
\end{equation}
where $\bfq^{\text{sub}}_t$ is a distribution over all substructures.

%First, we make an empirical study to explore show that most of the substructures in generated molecule are originally in the input molecule, as described in Table~\ref{table:oov}. 

However, the number of all possible substructures is usually quite large, for example, in Table~\ref{table:oov}, the vocabulary size of DRD2 dataset is 967, which means the number of categories is 967, which makes prediction more challenging especially for the rare substructures. 
%According to statistics in Table~\ref{table:oov}, more than 80\% substructures are kept in targeted molecule. Thus, we seek an extractive fashion for $\bfq_t^{\text{sub}}$. 

Inspired by pointer network~\cite{vinyals2015pointer,see2017get}  we design a similar strategy to copy some of the input sequence to the output. 
However, pointer network does not handle the case where the target molecule contains \textbf{Out-of-Input} (\textbf{\oov}) substructures, i.e., a novel substructure is not part of the input molecule. 
Borrowing the idea from sequence-to-sequence model~\cite{gu2016incorporating,see2017get}, we design a method to predict the weight of generating novel \oov substructures. 

\paragraph{Refine with novel substructures}
First, we use context vector $\bfc_t^{\text{sub}}$ and embeddings of input molecule graph and its scaffolding tree to determine the weight of generating novel substructures in current step,  
\begin{equation}
\label{eqn:oov}
\begin{aligned}
w_t^{\text{\oov}} = g_6(\bfc_t^{\text{sub}}, \bfz ),
\end{aligned}
\end{equation}
where $g_6(\cdot)$ is a fully-connected neural network with sigmoid activation. 
Thus, the weight ranges from 0 to 1. 
$w_t^{\text{\oov}}$ represents the probability that the model generate \oov substructure at $t$-th step. 
We assume that the weight depends on not only the input molecule (global information) and the current position in the decoder (local information). 
We use a representation $\bfz$ to express the global information of input molecule,   
\begin{equation}
\label{eqn:global}
\begin{aligned}
\bfz = \bigg[ \frac{1}{\vert\{\bfx_i^{\calT}\} \vert} \sum_i \bfx_i^{\calT}, \frac{1}{\vert\{\bfx_j^{G}\} \vert} \sum_j \bfx_j^{G} \bigg]
\end{aligned}
\end{equation}
where $\bfz$ is the concatenation of average embedding of all the scaffolding tree node and average embedding of all the graph node. 
Local information is represented by context vector $\bfc_t^{\text{sub}}$ computed via attention mechanism.

\begin{figure}[]
\centering
\subfigure[DRD2]{
\includegraphics[width=0.47\columnwidth]{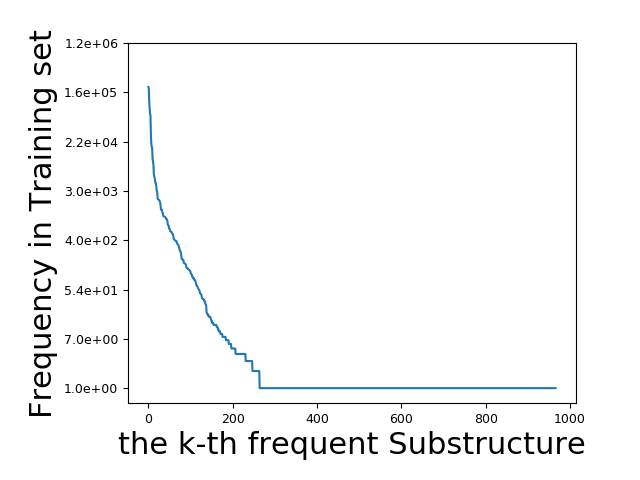}}
\subfigure[QED]{
\includegraphics[width=0.47\columnwidth]{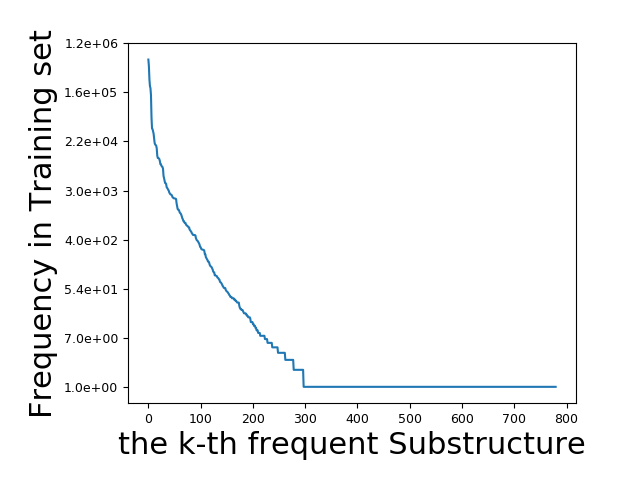}}
\caption{The frequency of different substructures on DRD2(a) and QED datasets(b). It indicates that the distribution of various substructures is highly imbalanced. 
}
\label{fig:distribution}
\end{figure}

\paragraph{Copy existing substructures}
After obtaining the weight of \oov substructure, we consider if and what  substructures to copy from input molecule. 
Each substructure in input molecule has an attention weight (normalized, so the sum is 1), which measure the contribution of the substructure to decoder.  
Then we use it to represent the selection probability for each substructure. 
Specifically, we define a sparse vector $\bfa$ as 
\begin{equation}
\label{eqn:select_prob}
\begin{aligned}
\{\bfa\}_{i} = \left\{
\begin{array}{ll}
 \sum_{j \in \mathcal{C} } \bfalpha_{j}^{\calT},& \mathcal{C}=\{j|\text{$j$-th node is i-th substruct}\},  \\
 0,&\text{i-th substructure not in $\calT$}, \\
\end{array}
\right.
\end{aligned}
\end{equation} 
where $ \bfa \in\mathbb{R}^{\vert S \vert}$, $\vert S\vert$ is size of   $\{\bfa\}_{i}$ represent $i$-th element of $\bfa$.  
Thanks to the normalization of attention weight (Equation~\ref{eqn:attentionweight}) $\{\bfalpha^{\calT}_1, \bfalpha^{\calT}_2, \cdots \}$, $\bfa$ is also normalized.  

The prediction at $t$-step is formulated as a hybrid of  
\begin{equation}
\label{eqn:hybrid}
\begin{aligned}
\tilde{\bfq}_{t}^{\text{sub}} = w_{t}^{\text{\oov}} \bfq_{t}^{\text{sub}} + (1 - w_{t}^{\text{\oov}}) \bfa_t.  
\end{aligned}
\end{equation} 
where $w_{t}^{\text{\oov}}$ balances the contributions of two distributions at $t$-th step.
If novel substructures are generated, we select the substructure from all substructures according to distribution $\bfq_t^{\text{sub}}$). 
Otherwise, we use the pointer network~\cite{vinyals2015pointer} to copy a certain substructure from the input molecule. The selection criteria for substructures in input molecule is proportional to the attention weight $\{\bfalpha^{\calT}\}$. 
%The neural architecture is shown in Figure~\ref{fig:copy}. 

\subsubsection{B. Graph decoder (Finetuning)} Finally, we describe graph decoder. The goal is to assemble nodes in scaffolding tree together into the correct molecular graph~\cite{jin2018junction}, as shown in Figure~\ref{fig:pipeline}. During learning procedure, all candidate molecular structures $\{G_i\}$ are enumerated and it is cast as a classification problem whose target is to maximize the scoring function of the right structure $G_o$
\begin{equation}
\label{eqn:decode3}
\begin{aligned}
& \mathcal{L}_g =  f^a(G_o) - \log \sum\limits_{G_i} \exp(f^a(G_i)),
\end{aligned}
\end{equation}
where $f^a(G_i) = \mathbf{h}_{G_i} \cdot \bfz_{G_o}$ is a scoring function that measure the likelihood of the current structure $G_i$, $\bfz_{G_o}$ is an embedding of graph $G_o$. 
On the other hand, when sampling, graph decoder is to enumerate all possible combinations, and pick the most likely structure in a greedy manner. 

\subsection{Adversarial Learning}
Adversarial training is used to further improve the performance, where the entire encoder-decoder architecture is regarded as the generator $G(\cdot) $, the target molecule $Y$ is regarded as the real sample, discriminator $D(\cdot)$ is used to distinguish real molecules from generated molecules by our decoder~\cite{jin2019learning}. Following \cite{jin2019learning}, $D(\cdot)$ is a multi-layer feedforward network.

\begin{table*}[tb]
\centering
\small
\caption{Empirical results measured by \textbf{Similarity} for various methods on different datasets. }
\label{table:similarity}
\begin{tabular}{ccccc|cccc}
\toprule[1pt]
\multirow{2}{*}{Method} & \multicolumn{4}{c}{Test Set} & \multicolumn{4}{c}{Test subset with infrequent substructures}  \\ 
& QED & DRD2  & LogP04 & LogP06  & QED & DRD2  & LogP04 & LogP06 \\
\hline
JTVAE & .2988 & .2997 & .2853 & .4643 & .2519 & .2634 & .2732 & .4238 \\ 
GCPN & .3081 & .3092 & .3602 & .4282 & .2691 & .2759 & .2973 & .3709  \\ 
Graph-to-Graph & .3145 & .3164 & .3579 & .5256 & .2723 & .2760 & .2901 & .4744  \\
\mname & {\bf .3211} & {\bf .3334} & {\bf .3695} & {\bf .6386} & {\bf .2982} & {\bf .3021} & {\bf .3234} & {\bf .5839}\\ 
\bottomrule[1pt]
\end{tabular}
\end{table*}

\begin{table*}[tb]
\small 
\centering 
\caption{Empirical results measured by \textbf{Property(Y)} for various methods on different datasets. }
\label{table:property}
\begin{tabular}{ccccc|cccc}
\toprule[1pt]
\multirow{2}{*}{Method} & \multicolumn{4}{c}{Test Set} & \multicolumn{4}{c}{Test subset with infrequent substructures}  \\ 
& QED & DRD2  & LogP04 & LogP06  & QED & DRD2  & LogP04 & LogP06 \\
\hline
JTVAE & .8041 & .7077 & 2.5328 & 1.0323 & .7292 & .6292 & 2.0219 & .7832  \\ 
GCPN & .8772 & .7512 & 3.0483 & {\bf 2.148} & .7627 & .6743 & 2.5413 & 1.813 \\ 
Graph-to-Graph & .8803 & .7641 & 2.9728 & 1.983 & .7632 & .6843 & 2.4083 & 1.778  \\ 
\mname & {\bf .8952} & {\bf .7694} & {\bf 3.1053} & {2.021} & {\bf .7899} & {\bf .7193} & {\bf 2.7391} &  {\bf 1.820} \\ 
\bottomrule[1pt]
\end{tabular}
\end{table*}

\begin{table*}[tb]
\centering
\caption{Empirical results measured by \textbf{SR1 (Success Rate)} for various methods on different dataset. For QED and DRD2, regarding SR1, when the similarity between input and generated molecule ($\lambda_1$) is greater than $0.3$ and property of generated molecule ($\lambda_2$) is greater than $0.6$, we regard it ``success''. For LogP04 and LogP06,  $\lambda_1 = 0.4$ and  $\lambda_2 = 0.8$}
\label{table:sr}
\begin{tabular}{ccccc|cccc}
\toprule[1pt]
\multirow{2}{*}{Method} & \multicolumn{4}{c}{Test Set} & \multicolumn{4}{c}{Test subset with infrequent substructures}  \\ 
& QED & DRD2  & LogP04 & LogP06  & QED & DRD2  & LogP04 & LogP06 \\
\hline
JTVAE & 43.32\% & 34.83\% & 38.43\% & 43.54\%  & 38.91\% & 29.32\% & 35.32\% & 40.43\%  \\ 
GCPN & 47.71\% & 44.05\% & 56.43\% & 52.82\% & 42.80\% & 37.82\% & 42.81\% & 43.29\% \\ 
Graph-to-Graph & 48.16\% & 45.73\% & 56.24\% & 55.15\% & 43.43\% & 38.39\% & 42.83\% & 47.02\%  \\ 
\mname & {\bf 50.26\%} & {\bf 47.91}\% & {\bf 56.47\%} & {\bf 57.64\%} & {\bf 47.82\%}  & {\bf 42.72\%} & {\bf 45.01\%} & {\bf 50.05\%}  \\ 
\bottomrule[1pt]
\end{tabular}
\end{table*}

\begin{table*}[tb]
\centering
\caption{Empirical results measured by \textbf{SR2 (Success Rate)} for various methods on different dataset. For QED and DRD2, regarding SR2, when the similarity between input and generated molecule is greater than $0.4$ ($\lambda_1$) and property of generated molecule is greater than $0.8$ ($\lambda_2$), we regard it ``success''. For LogP04 and LogP06, $\lambda_1 = 0.4$ and  $\lambda_2 = 1.2$. }
\label{table:sr2}
\begin{tabular}{ccccc|cccc}
\toprule[1pt]
\multirow{2}{*}{Method} & \multicolumn{4}{c}{Test Set} & \multicolumn{4}{c}{Test subset with infrequent substructures}  \\ 
& QED & DRD2  & LogP04 & LogP06  & QED & DRD2  & LogP04 & LogP06 \\
\hline
JTVAE & 20.72\% & 9.13\% & 21.32\% & 18.32\% & 17.53\% & 7.18\% & 19.93\% &17.16\% \\ 
GCPN & 23.08\% & 14.94\% & 27.01\% & 25.29\% & 18.98\% & 12.64\% & 23.81\% & 23.02\% \\ 
Graph-to-Graph & 24.34\% & 15.31\% & 26.95\% & 25.30\% & 20.82\% & 12.88\% & 23.53\% & 23.82\%\\ 
\mname & {\bf 27.23\%} & {\bf 17.31\%} & {\bf 27.88\%} & {\bf 26.58\%} & {\bf 25.32\%} & {\bf 15.26\%} & {\bf 25.62\%} & {\bf 25.91\%} \\ 
\bottomrule[1pt]
\end{tabular}
\end{table*}

\iffalse
\begin{table*}[]
\centering
\caption{Accuracy comparison between frequent and infrequent substructure on QED dataset. The value is averaged over either frequent or infrequent substructures. }
\label{table:infrequent}
\begin{tabular}{ccccccc}
\toprule[1pt]
\multirow{2}{*}{Method} & \multicolumn{3}{c}{Frequent Substructure} & \multicolumn{3}{c}{Infrequent Substructure}  \\ 
& Recall &  Precision & F-score & Recall &  Precision & F-score  \\
\hline
JTVAE & $0.7147 \pm 0.0897$ & $0.3601 \pm 0.2897$ & $0.4589 \pm 0.2560$ & $0.0434 \pm 0.0614$ & $0.0601 \pm 0.0937$  & $0.0517 \pm 0.1056$  \\ 
Graph-to-Graph & $0.8050 \pm 0.0640$ & $0.4201 \pm 0.2732$ & $0.5508 \pm 0.2381$ & $0.0479 \pm 0.0503$ & $0.0678 \pm 0.0957$  & $0.0603 \pm 0.0738$  \\ 
\mname & $0.8091 \pm 0.0671$ & $0.4300 \pm 0.2873$ & $0.5603 \pm 0.2428$ & $0.1782 \pm 0.0473$ & $0.1296 \pm 0.0957$  & $0.1539 \pm 0.0735$  \\ 
\bottomrule[1pt]
\end{tabular}
\end{table*}
\fi

\section{Experiment}
%In this section, we describe our empirical procedures. We start with the description of experimental setup, including dataset description, preprocessing procedure, baseline methods and evaluation metrics. 
In the experiment section, we want to answer the following questions:
\begin{itemize}
    \item {\bf Q1:} Can \mname generate better molecules than other graph-based methods?
%    \item {\bf Q2:} Why do we focus on infrequent substructure?
    \item {\bf Q2:} How well does \mname handle input molecules with rare substructures?
\end{itemize}

\subsection{Molecular Data} 
First, we introduce the molecule data that we are using. 
ZINC contains 250K drug molecules extracted from the ZINC database
% \footnote{The ZINC database is a curated collection of commercially available chemical compounds prepared especially for virtual screening, \url{https://zinc.docking.org/}}
~\cite{sterling2015zinc}. 
We extract data pairs from ZINC, which will be described later.  We list the basic statistics for the datasets in Table~\ref{table:statistics}.

\subsection{Molecular Properties}\ \\ 
In drug discovery, some properties are crucial in evaluating the effectiveness of generated drugs. In this paper, following~\cite{jin2019learning}, we mainly focus on following three properties. 
\begin{itemize}
\item \textbf{Dopamine Receptor (DRD2)}. DRD2 score is to measure a molecule’s biological activity against a biological target named the dopamine type 2 receptor (DRD2). DRD2 score ranges from 0 to 1.
\item \textbf{Drug likeness (QED)}~\cite{bickerton2012quantifying}. QED score is an indicator of drug-likeness ranging from 0 to 1.  
\item \textbf{Penalized LogP}. Penalized logP is a logP score that accounts for ring size and synthetic accessibility~\cite{ertl2009estimation}.
\end{itemize}
For each SMILES string in ZINC, we generate the QED, DRD2 and LogP scores using Rdkit package~\cite{landrum2006rdkit}. 
For all these three scores, higher is better. Thus, for the training data pairs $(X,Y)$, $X$ is the input molecule with lower scores while $Y$ is a generated molecule with higher scores based on $X$. 

\subsection{Generation of Training Pairs}

To construct training data set, we find the molecule pair $(X,Y)$ following ~\cite{jin2019learning}, where $X$ is the input molecule and $Y$ is the target molecule with desired property. Both $X$ and $Y$ are from the whole dataset and have to satisfy two rules: (1) they are similar enough, i.e., $\text{sim}(X,Y)\geq \eta_1$; (2) $Y$ has significant property improvement over $X$, i.e., $\text{property}(Y) - \text{property}(X) \geq \eta_2$, $\text{property}(\cdot)$ can be DRD2, QED, LogP score as mentioned above. 
$\eta_1 = 0.4$ for LogP04 while $\eta_1 = 0.6$ for LogP06. We use the public dataset in \cite{jin2019learning} with paired data.

\paragraph{Infrequent Substructures}
We pay special attention on the infrequent substructure. Based on observation from empirical studies (for example in Figure~\ref{fig:distribution}), regarding a substructure if it occurs less than 2,000 times in training set, we call it ``infrequent substructures``. Otherwise, we call it ``frequent substructure''.

\subsection{Baseline Methods}\ \\ 
We compare our method with some important baseline methods, which represents state-of-the-art methods on this task.  
\begin{itemize}
%\item \textbf{JTVAE}~\cite{jin2018junction}. Our method is developed on basis of JTVAE. The difference is that JTVAE is totally unsupervised while our method uses property score as the guidance. 
% \item \textbf{Seq2Seq}~\cite{sutskever2014sequence}. If we use SMILES string to represent molecule, we can apply Sequence-to-Sequence model to training the molecule pairs. 
\item \textbf{JTVAE}~\cite{jin2018junction}. scaffolding tree variational autoencoder (JTVAE) is a deep generative model that learns latent space to generate desired molecule. It also uses encoder-decoder architecture on both scaffolding tree and graph levels. 
\item \textbf{Graph-to-Graph}~\cite{jin2019learning}. It is the most important benchmark method as described above.
\item \textbf{GCPN}~\cite{You2018-xh} uses graph convolutional policy networks to generate molecular structures with specific properties. It exhibits state-of-the-art performance in RL-based method.
\end{itemize}
Note that we also tried Sequence-to-Sequence model~\cite{sutskever2014sequence} on SMILES strings, but the resulting model generates too many invalid SMILES strings to compare with all the other graph-based methods. This further confirmed that graph generation is a more effective strategy for molecular optimization.

\subsection{Evaluation}  
During evaluation procedure, we mainly focus on several evaluation metrics, where $X$ is the input molecule in test set, $Y$ is the generated molecule. 
\begin{itemize}
\item \textbf{Similarity}. We evaluate the molecular similarity between the input molecule and the generated molecule, measured by Tanimoto similarity over Morgan fingerprints~\cite{rogers2010extended}. The similarity between molecule $X$ and $Y$ is denoted $\text{sim}(X,Y)$, ranging from 0 to 1. 
\item \textbf{Property of generated Molecules}. The second metric is the property scores of generated molecules. It is defined as $\text{Property}$(Y), where property could be  including QED-score, DRD2-score and LogP-score, evaluated using Rdkit package~\cite{landrum2006rdkit}.  
\item \textbf{Success Rate (SR)}. Success Rate is a metrics that consider both similarity and property improvement. Since the task is to generate a molecule that (i) is similar to input molecule and (ii) have property improvement at the same time. We design a criteria to judge whether it satisfied these two requirement: (a) Input and generated molecules are similar enough, $\text{sim}(X,Y)\geq \lambda_1$; (b) improvement are big enough, i.e., $\text{property}(Y)  \geq \lambda_2$. The selection of $\lambda_1$ and $\lambda_2$ depend on datasets. 
%\item \textbf{Novelty}. The ultimate goal of drug discovery is to find new molecule with desired property. So we want to generate as much new molecule as possible. If the generated molecule doesn't occur in training set and is different from input molecule in test set, we regard it as a new molecule. 
%\item \textbf{Run time \& Model Size}. We also report run time (for training) and the model size. The unit of run time is hour. 
\end{itemize}
Among these metrics, similarity and property improvement are the most basic metrics. 
For all these metrics except run time and model size, higher values are better. 

\subsection{Implementation Details}
\label{sec:details}

In this section, we provide the implementation details for reproducibility, especially the setting of hyperparameters. 
We follow most of the hyperparameter setting of \cite{jin2019learning}. 
For all these baseline methods and datasets, maximal epoch number is set to 10, batch size is set to 32. 
During encoder module, embedding size is set to 300. The depth of message passing network are set to 6 and 3 for tree and graph, respectively. 
The initial learning rate is set to $1e^{-3}$ with the Adam optimizer. 
Every epoch learning rate is annealed by 0.8. 
We save the checkpoint every epoch during training procedure. When evaluating, from all the checkpoints, we choose the one that achieves highest success rate (SR1) on validation set as the best model and use it to evaluate the test set. 
During adversarial training, discriminator $D(\cdot) $ is a three-layer feed-forward network with hidden layer dimension 300 and LeakyReLU activation function. 
For all these datasets, model size of \mname is around 4M, nearly same as Graph-to-Graph model. The code is available\footnote{\url{https://github.com/futianfan/CORE}}.

\subsection{Results}
\label{sec:results}

The results for various metrics (including similarity, property improvement, success rate) are shown in Table~\ref{table:similarity}, \ref{table:property}, \ref{table:sr}, \ref{table:sr2}, respectively. 
Now we can answer three questions proposed in the beginning of the section. 
\begin{itemize}
    \item {\bf Q1 Performance on all molecules}: We compare \mname with baseline methods on complete test sets. \mname outperforms baselines in all the measures.
Specifically, when measured by success rates SR (both SR1 and SR2, Table~\ref{table:sr} and~\ref{table:sr2}), \mname obtained about 2\% absolutely improvement over the best baseline. 
When measured by SR2, it can achieve over 10\% relatively improvement on QED and DRD2. 
\item {\bf Q2 Performance on molecules with infrequent substructures}: 
 Test subset with infrequent substructures is more challenging, because for all the methods the performance would degrade on infrequent subset. Thus, it is worth to pay special attention to the molecule with infrequent substructure. 
When measured on the test subset with infrequent substructure, \mname achieves more significant improvement compared with the complete test set. Concretely, \mname achieves 21\% and 18\% relatively improvement in success rate (SR2) in QED and DRD2, and more than 3\% absolutely improvement in SR (both SR1 and SR2). 
In a word, \mname gain more improvement in rare substructure compared with the whole test set. 
\end{itemize}

\section{Conclusion}

In this paper, we propose a deep generative model for creating molecules with more desirable properties than the input molecules. 
The state-of-the-art Graph-to-Graph methods grow the new molecule structure by iteratively predicting substructures from a large set of substructures, which is challenging especially for infrequent substructures.  
To address this challenge, we have propose a new generating strategy called ``\texttt{Co}py\&\texttt{Re}fine'' (\mname), where at each step the generator first decides whether to copy an existing substructure from input $X$ or to generate a new substructure from the large set. 
The resulting \mname mechanism can significantly outperform several latest molecule optimization baselines in various measures, especially on rare substructure. 

\section*{Acknowledgement}
This work was in part supported by the National Science Foundation award IIS-1418511, CCF-1533768 and IIS-1838042, the National Institute of Health award NIH R01 1R01NS107291-01 and R56HL138415.

\bibliographystyle{aaai}
\bibliography{ref}

\end{document}